\documentclass[conference]{IEEEtran}

  	\usepackage[pdftex]{graphicx}
  	\graphicspath{{../pdf/}{../jpeg/}}
	\DeclareGraphicsExtensions{.pdf,.jpeg,.png}

	\usepackage[cmex10]{amsmath}
	\usepackage{mathabx}
	\usepackage{algorithmic}
	\usepackage{array}
	\usepackage{mdwmath}
	\usepackage{mdwtab}
	\usepackage{eqparbox}
	\usepackage{url}
    \usepackage[utf8]{inputenc}
    \usepackage{siunitx}
    \usepackage{multirow}
\begin{document}

\title{Capsule Neural Network based Height Classification using Low-Cost Automotive Ultrasonic Sensors}

 \author{Maximilian Pöpperl, Raghavendra Gulagundi, Senthil Yogamani and Stefan Milz\\Valeo Schalter und Sensoren GmbH\\Kronach, Germany\\maximilian.poepperl@valeo.com}

\maketitle

\begin{abstract}
High performance ultrasonic sensor hardware is mainly used in medical applications. Although, the development in automotive scenarios is towards autonomous driving, the ultrasonic sensor hardware still stays low-cost and low-performance, respectively. To overcome the strict hardware limitations, we propose to use capsule neural networks. By the high classification capability of this network architecture, we can achieve outstanding results for performing a detailed height analysis of detected objects. We apply a novel resorting and reshaping method to feed the neural network with ultrasonic data. This increases classification performance and computation speed. We tested the approach under different environmental conditions to verify that the proposed method is working independent of external parameters that is needed for autonomous driving.
\end{abstract}

\IEEEoverridecommandlockouts

\IEEEpeerreviewmaketitle


\section{Introduction}
Ultrasound is a relatively old technology used in automotive applications. Starting from simple warning systems for parking, ultrasonic sensors evolved rapidly to an essential component of modern driving assistance \cite{us_parking}. One of the key points of the high popularity of these sensors is for sure the low production costs. Although this increases the spread of ultrasonic technology, the focus in the ultrasonic sensor development for automotive applications has mainly been on cost reduction rather than on performance improvements \cite{advanced_us}.\\
To keep pace with the upcoming requirements for autonomous driving, ultrasonic systems have a high need for new developments at the moment. However, major hardware changes are prevented by existing standards and automotive low-cost expectations. Consequently, another starting point for improvements needs to be found. \\
A role model to solve the performance issues are medical ultrasonic applications, such as prenatal diagnosis or screening \cite{prenatal_us}. In contrast to automotive setups, the sensor hardware is much more complex and costly. Nevertheless, some processing approaches are adaptable to automotive scenarios. Especially, the methods used for pattern recognition are of interest. Whereas in automotive applications algorithms are mainly based on thresholds, medical ones use classification methods like adaptive boosting or support vector machine \cite{pattern_recog}.\\
In addition, deep learning based classification is a highly frequented topic in medical ultrasonic diagnosis. The applied approaches cover different network architectures including multilayer perceptrons (MLP) \cite{us_mlp}, convolutional neural networks (CNN) \cite{us_cnn} as well as recurrent neural networks (RNN) \cite{us_rnn}. The applications are as manifold as the network architectures and last from pattern detection and tracking to segmentation \cite{us_segmentation} or image reconstruction \cite{us_image_rec}.\\
Although approaches like image reconstruction and tracking are thinkable using automotive ultrasonic sensors, their usage is prevented by low-performance processing units in present cars. Consequently, we need to improve the functionality of ultrasonic sensors by a classification of detected objects. Basically different object properties can be determined with a single sensor, e.g. width, material or height. As the object height is most important for driving assistance systems, we use it exemplary to evaluate our proposed classification method.
\begin{figure}[b]
\centering{\includegraphics[width=68mm]{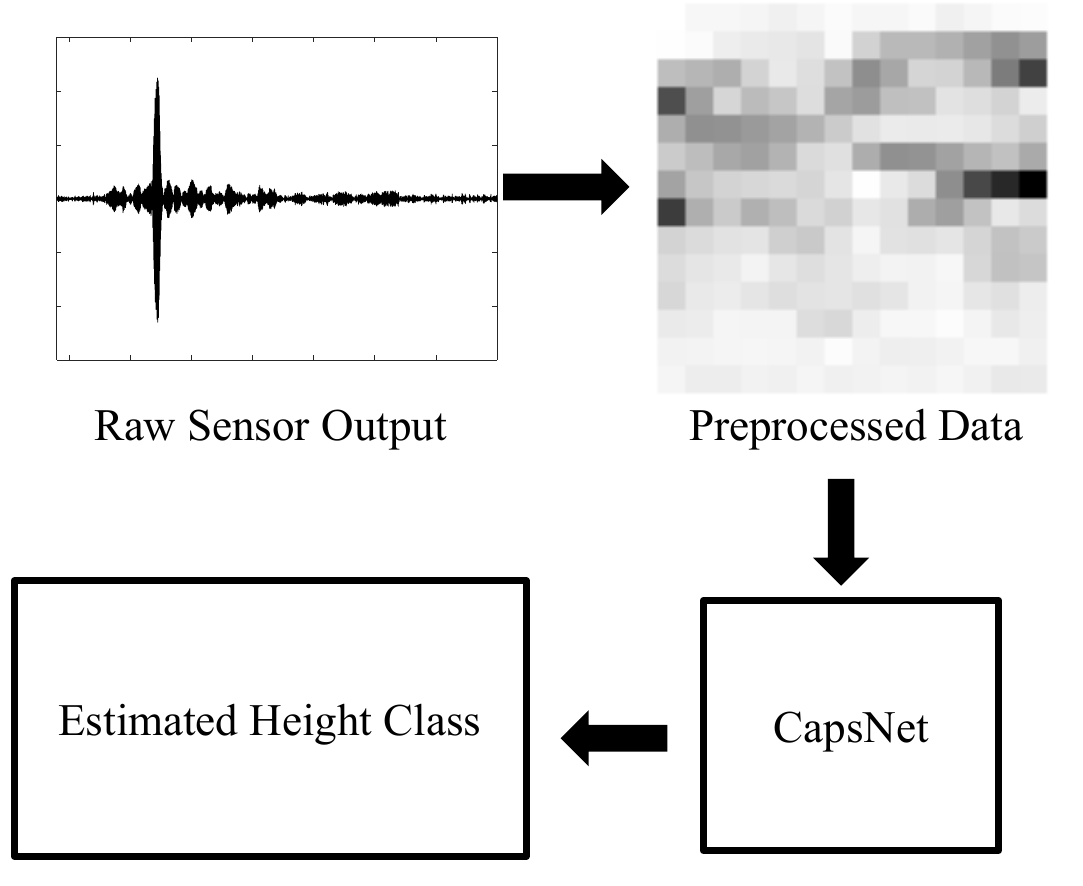}}
\caption{Preprocessing and classifications steps of the proposed approach}
\label{approach}
\end{figure}\\
The low-cost sensor and processing unit additionally restrict the applicable network architectures to CNNs or similar computational efficient implementations. Due to the low performance of a sensor, the requirements for a classification algorithm are very challenging. Thus, network architecture like capsule neural networks (CapsNet) \cite{capsnet_paper1}, \cite{capsnet_paper2} that achieve improved performance by similar computational effort as CNNs are suited well to solve our ultrasonic classification problem.\\
Subsequent to the description of our measurement setup, we propose a special preprocessing and data preparation method. This method is necessary to feed the neural network adequately, but also to increase classification accuracy and speed. The neural network basically follows the capsule approach from \cite{capsnet_paper1}. We adapted it to the demands of ultrasonic data. An overview of the method is given in Fig.\:\ref{approach}. At the end we verify our results and show the independence of environmental influences of the proposed height estimation.


\section{Measurement setup}
\begin{figure}[b]
\centering{\includegraphics[width=65mm]{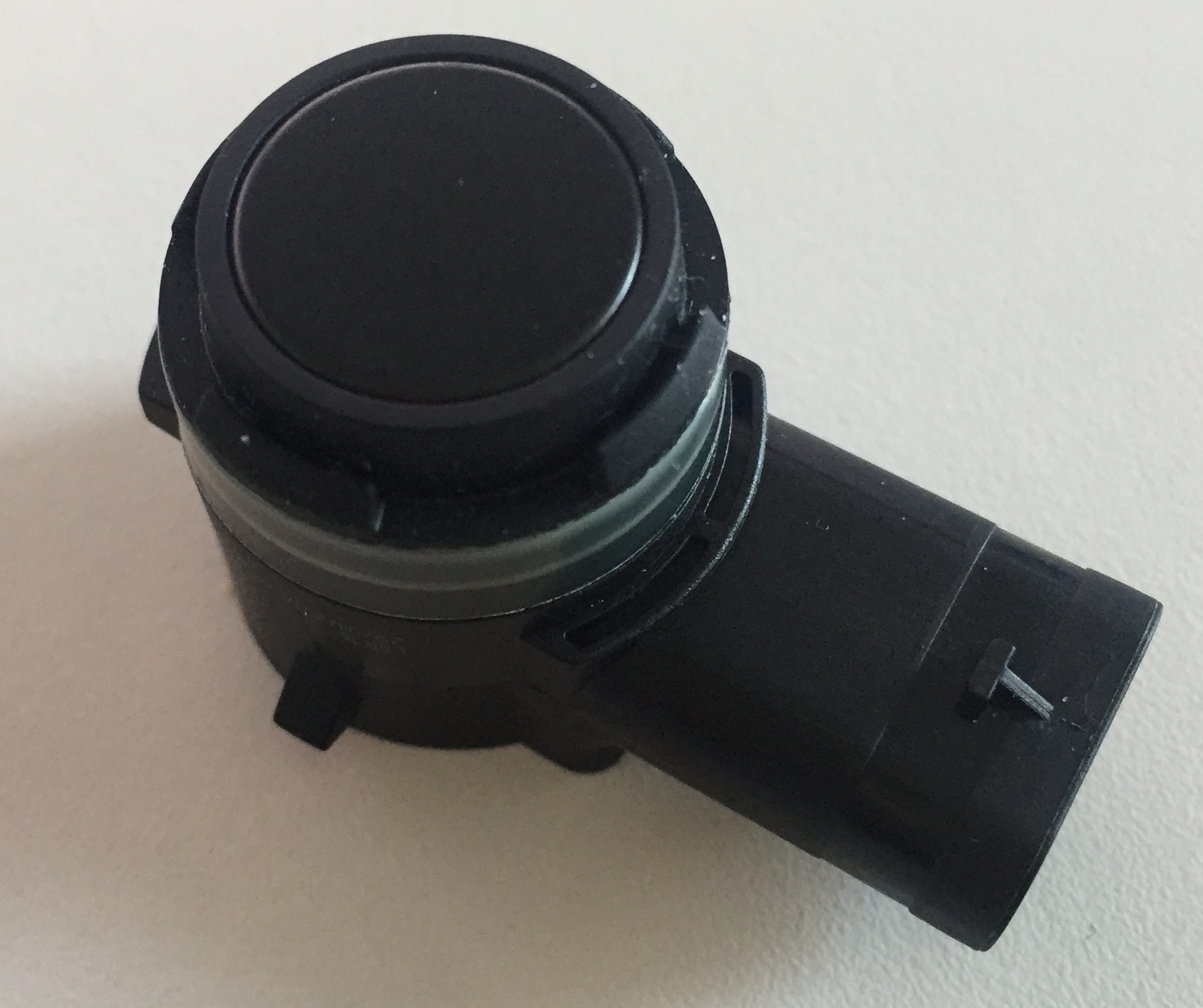}}
\caption{Valeo ultrasonic sensor used for measurements and classification}
\label{sensor}
\end{figure}

A conventional automotive ultrasonic sensor is used. The sensor has a single membrane. The transmit signal is a modulated pulse at a center frequency of \SI{51.2}{\kilo\hertz}. The bandwidth of the pulse is about \SI{3}{\kilo\hertz}. The sensor is shown in Fig.\:\ref{sensor}. We use a monostatic measurement setup, i.e. the transmitting sensor is also used to receive the reflected signals. The sensor provides raw data from the piezoelectric element. An analog filter bank is used for signal conditioning before the signal is converted from analog to digital. Further processing steps are done on a computer and are described in detail in section\:\ref{preprocess}. \\
In order to verify that the proposed algorithm is independent of environmental influences, which is very important in automotive applications due to changing conditions, we recorded a wide variety of scenarios. Beside dry and wet ground, we also measured different ground types, such as asphalt, gravel and grass. The temperature in our measurements has varied between \SI{5}{\celsius} to \SI{25}{\celsius}, so that temperature dependent effects of the membrane are covered.\\
Furthermore, we use different objects, e.g. curbs, tubes and walls. The objects itself also have different heights, as for example some curbs can be driven over without braking or at least require a reduced speed. We found that the object distance and lateral position to the sensor also has a significant influence on the height classification. Hence, we extended the data set by considering object distances between \SI{0.5}{\meter} and \SI{2.5}{\meter}.\\
In total our dataset contains 21,600 measured responses from a single sensor. On the one hand this amount of data is needed to train the neural network adequately. But on the other hand it is also necessary in order to cover all environmental influences and achieve high robustness that is required in driving assistance applications.
We used a manual labeling as we only consider static scenarios. Anyway, we need to generate each setup manually. The reason for this is that typically high and very low objects are frequent in automotive scenarios. Objects in between very low and high heights are rare, so that they need to be placed artificially. The data is labeled with respect to four different height classes that are chosen according to automotive specification. The first height is chosen below \SI{10}{\centi\meter}, so that it is possible to drive over. The next class is defined between \SI{10}{\centi\meter} and \SI{30}{\centi\meter}. For objects in this height, it is possible to drive over without damage using a suitable car and reduced speed. The third class contains objects with a height between \SI{30}{\centi\meter} and \SI{50}{\centi\meter}, where driving over without damage is not possible anymore. A door can still be opened. The fourth class includes all objects that are higher than \SI{50}{\centi\meter}. Again driving over is not possible. Furthermore a door cannot be opened anymore, so that more space between the car and such an object is required when parking. \\
Subsequent to the data acquisition and labeling, the data needs to be preprocessed in a way that allows the feeding into the proposed neural network and optimizes performance.


\section{Preprocessing and Data Preparation} \label{preprocess}
The output of the ultrasonic sensor is just the voltage of the piezoelectric element. After an active filter cascade, the data is converted to digital. This allows us to analyze different processing methods. We divided the processing into two different steps:
\begin{enumerate}
\item Signal conditioning
\item Reorganization
\end{enumerate}
As each part of the processing contains different processing steps, they are described in more detail in the following. Basically the signal conditioning is used to improve the ultrasonic signal itself, whereas the restructuring is applied to make the data suitable to be fed in a neural network.

\begin{figure}[tb]
\centering{\includegraphics[width=88mm]{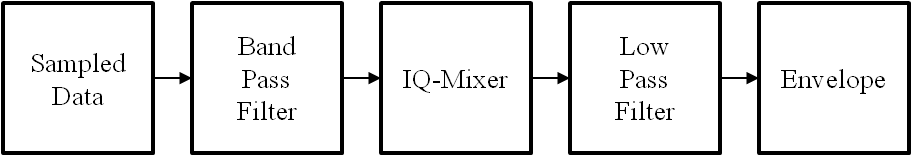}}
\caption{Signal conditioning steps}
\label{conditioning}
\end{figure}
\subsection{Signal conditioning}
The ultrasonic raw data $s_{r}$ is sampled with a sampling frequency of \SI{110}{\kilo\hertz} to meet the Nyquist sampling theorem. After the analog-to-digital conversion, a digital band-pass filter is used to reduce noise and to remove the DC offset that is necessary for the conversion. We use a high order finite impulse response filter $s_{b}$ in order to avoid non-linear phase distortion. The noise reduced signal $s_{p1}$ results from:
\begin{equation}
s_{p1} = s_{r}*s_{b}
\end{equation}
The signal is converted to the complex baseband afterwards. This is done in two steps. First we use an IQ-mixer to remove the carrier frequency $f_c$ of the signal:
\begin{equation}
s_{p2} = s_{p1}\cdot\exp[-2j\pi f_c t]
\end{equation}
Subsequently, we suppress image frequencies by a low-pass filter with the filter response $s_{l}$. The output of the filter is:
\begin{equation}
s_{p} = s_{p2}*s_{l}
\end{equation}
This is in fact the complex envelope of the ultrasonic raw signal. As the filters are adapted to the transmit signal, the down conversion of the signal can be done without any loss of information, i.e. all information of the signal that is available in the raw data is also available in the complex envelope \cite{digital_comm}. 
The loss-free implementation is essential for the following classification, as the proposed method avoids any reduction of the classification accuracy in this way. The signal conditioning is summarized in Fig.\:\ref{conditioning}.\\

\subsection{Reorganization}
\begin{figure}[b]
\centering{\includegraphics[width=88mm]{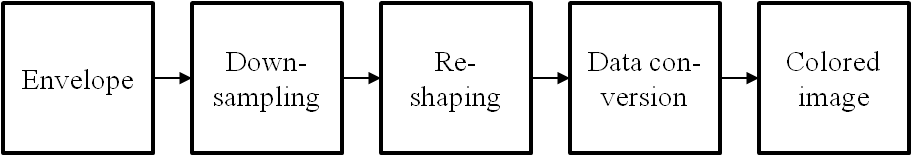}}
\caption{Reorganization steps}
\label{reorganization}
\end{figure}

\begin{figure}[b]
\centering{\includegraphics[width=30mm]{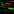}}
\caption{Complex image for network input (red channel: real part, green channel: imaginary part)}
\label{cpx_data}
\end{figure}

The previous processing was necessary to remove noise and other unwanted disturbances from the signal. The aim of the reorganization procedure is completely different. Beside the reformatting of the data, so that it can be fed to the neural network, we compress the envelope data and optimize it for classification speed. An overview over the reorganization process is given in Fig.\:\ref{reorganization}.\\
In a first step we perform a down-sampling to reduce the number of data points that need to be handled by the neural network. Although the signal is in complex baseband, the sampling frequency of the signal was not changed so far. It is still \SI{110}{\kilo\hertz}. Using a measurement duration of \SI{30}{\milli\second}, 3,300 measurement points per measurement are available. After the conversion to the complex baseband, the highest available frequency is given by the bandwidth of the signal that is \SI{3}{\kilo\hertz}. Consequently, we can reduce the number of data points drastically. To avoid any loss of information even with varying temperature that changes the resonance frequency of the membrane, we choose a sampling frequency of \SI{6.5}{\kilo\hertz}. This reduces the number data points of the envelope to 196. The number of input neurons for the neural networks is reduced by a factor of 16.8. Additionally the number of calculations in deeper layers decreases, so that a significant improvement in the classification speed is achieved.\\
To further optimize the prediction speed, we rearrange the ultrasonic data. The ultrasonic data that is of dimension $196 \times 1$ is reshaped to a complex data array of dimension $14 \times 14$. The two dimensional shape allows the use of two dimensional convolutional kernels and a two dimensional pooling that reduces computational complexity.\\
To apply methods from image classification, where CNNs and CapsNets are mainly used, the ultrasonic data is rescaled and quantized to \SI{16}{bits}. As the analog to digital conversion is done using \SI{12}{bits}, information is not lost. The advantage is that the data can be stored in an suitable graphics format, such as portal network graphics.\\
Principally, it is possible to just store the absolute values of the envelope in a gray-scale graphic. We will show in section \ref{results} that this would lead to a degradation of the classification performance. Hence, real and imaginary part of the complex envelope need to be considered. As we use standard graphics format, the real and the imaginary parts can be saved in different color channels of a graphic. An example that uses the red and the green channel is shown in Fig.\:\ref{cpx_data}. The proposed method is additionally advantageous as we can use the channel representation in the neural network to handle the complex input data.\\
Summarizing, the preprocessing and data preparation is used to improve the signal quality of the raw signal by filtering and the conversion to the complex baseband. Downsampling and reshaping lead to a reduction of the processing time. The conversion of the complex ultrasonic data into a conventional graphics format using two color channels allows the application of existing network topologies that are mainly used in image processing, e.g. CapsNets.

\section{Classification Method}
\begin{figure*}[ht]
\centering{\includegraphics[width=180mm]{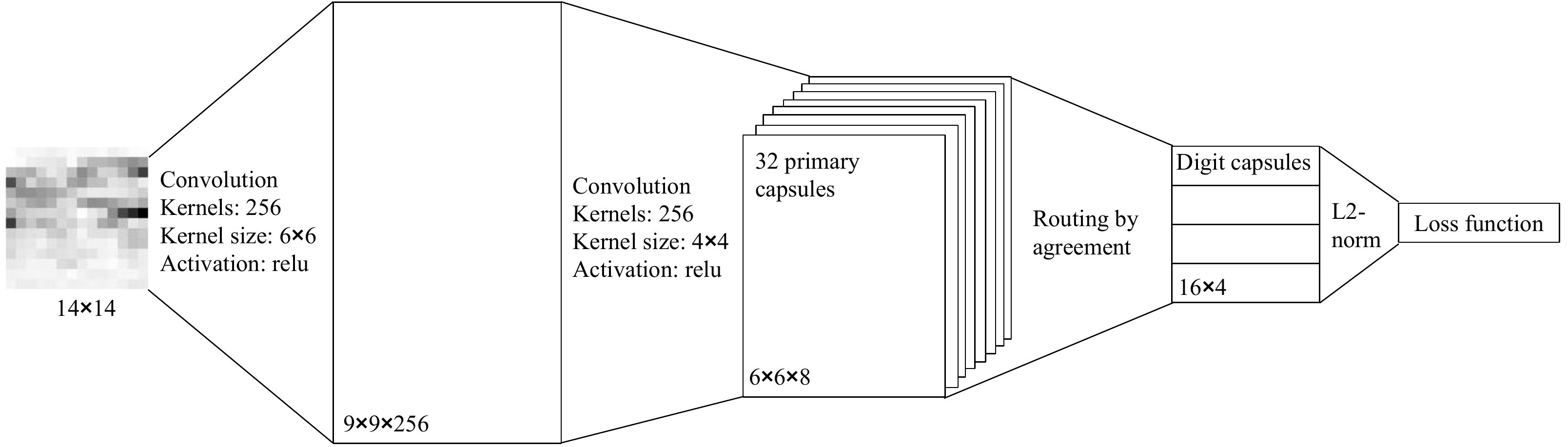}}
\caption{Architecture of CapsNet}
\label{capsnet_schem}
\end{figure*}
Due to the format of the processed ultrasonic data as two dimensional array with two channels, we can basically apply a wide variety of neural networks. However, a computational efficient implementation must be used in order to meet the requirements in automotive applications and to fit in standard processing units. Additionally, applying a two dimensional CNN assumes that there is a physical meaning in the vertical neighborhood of amplitude values in the reshaped data. As this is obviously not the case, the application of conventional CNNs is not effective.\\
CapsNets preserve the location of the amplitude values, so that merging physically not linked amplitude values is avoided. Furthermore CapsNets come up with an outstanding classification performance. This helps to compensate the low hardware performance of automotive ultrasonic systems. Altogether CapsNets turn out to be the best suited network topology to perform the height classification on our ultrasonic dataset. Due to our particular dataset, we need to adjust the method that is introduced in \cite{capsnet_paper1}. A detailed description of our used network is given in the following.\\
The network consists of four layers. The first layer is convolutional layer with 128 convolutional kernels. Each kernel is $6 \times 6$. The stride in this layer is one. We use a RELU activation function for computational efficiency. \\
The second layer again applies a convolutional kernel. We use a kernel of $4 \times 4$. Again the stride is chosen to 1 and the activation function is RELU. Additionally, to the convolutional step, the second layer reshapes and rearranges the data. The data is split into 32 separate parts. These parts are called primary capsules in the following. Each capsule has a dimension of $6 \times 6 \times 8$. \\
The capsules are weighted by specific weight matrices $\boldsymbol{W}_{ij}$, before the routing process is applied. The dimensions of the weight matrices are $8 \times 8$. The routing is the crucial step in CapsNets. We use a standard routing-by-agreement approach. This method is sufficient. Indeed, the more advanced EM-routing comes up with additional functionality that is on the one hand not needed in our application and on the other hand is computational more complex. Hence, we focus on the simpler method. The results of the routing are digit capsules. As there are four different height classes, we also get four digit capsules of dimension $1 \times 8$ each. \\
The loss of the network is calculated using a separate margin loss, as described in \cite{capsnet_paper1}:
\begin{equation}
\begin{split}
L_k=T_k\max(0,m^+-\Vert \boldsymbol{v}_k \Vert)^2+\\
\lambda(1-T_k)\max(0,\Vert \boldsymbol{v}_k \Vert - m^-)^2
\end{split}
\end{equation}
where $T_k$ is a flag if the sample is of class k and $\boldsymbol{v}_k$ are the digit capsules. $m^+$ and $m^-$ are the decision factors and $\lambda$ is a weighting factor. For the training procedure, the overall loss can be calculated from the sum of the individual losses.For the decision in the prediction, the maximum of the L2-norms of all digit capsules is decisive. \\
An illustrative summary of the proposed network architecture is given in Fig.\:\ref{capsnet_schem}. 

\begin{table}[b]
\centering
\caption{Confusion matrix of CapsNet for validation dataset}
\label{confmatrix}
\begin{tabular}{cc | c | c | c | c |}
\cline{3-6}
& & \multicolumn{4}{c|}{Actual}\\ \cline{3-6}
& & lowest & low & high & highest \\ \cline{1-6}
\multicolumn{1}{|c|}{\multirow{4}{*}{Predicted}} & lowest & 236 & 0 & 0 & 0 \\ \cline{2-6}
\multicolumn{1}{|c|}{\multirow{4}{*}{}} & low & 0 & 240 & 0 & 0 \\ \cline{2-6}
\multicolumn{1}{|c|}{\multirow{4}{*}{}} & high & 4 & 0 & 240 & 0 \\ \cline{2-6}
\multicolumn{1}{|c|}{\multirow{4}{*}{}} & highest & 0 & 0 & 0 & 240 \\ \cline{1-6}
\end{tabular}
\end{table}

\section{Implementation and validation} \label{results}
To train the neural network and to validate the proposed classification method, a dataset of 21,600 different measurements has been created. We equalize the number of measurements per class in training and validation data set avoid offsets in the classification and interpretation. The equalized dataset contains 4,800 shuffled measurements from the original dataset. We split the shuffled dataset into training and validation dataset. Due to the high amount of data, a holdout validation is used with 960 measurements. By covering different environmental conditions, we ensure the usability of the trained network for realistic driving scenarios.\\
We implement the network using Tensorflow. Adam optimizer is used to train the network. Stochastic batches with 100 elements are used to determine the weights and biases of the network.\\
At first, the network architecture is evaluated with the complex ultrasonic data. To illustrate the outstanding performance of the proposed approach, the confusion matrix is shown in Table\:\ref{confmatrix}. The matrix is based on the validation data that has 960 measurements. The classes are named: "lowest", "low", "high" and "highest". The calculated overall validation accuracy is 99.6\%. To evaluate the classification in detail, we additionally need to have a look on the confusion matrix. Our dataset contains 480 actual "low" and "lowest" objects. The predicted output of network has 476 objects that are of class "low" and "lowest". Hence, the network tends to predict low objects to high. As it is less bad stopping in front of a drivable object than driving against a non-drivable object, this tendency is acceptable for autonomous driving applications. Furthermore, the confusion matrix shows that only "lowest" objects are wrongly classified as "high". A possible solution to handle such cases is a confirmation of the object height by at least two measurements. This approach is applicable because of the outstanding accuracy. By considering two or more sequential measurements in a realistic scenario, the height classification achieves accuracies that are suitable for autonomous driving.\\ 
Autonomous driving additionally requires a real-time processing of the ultrasonic data. The prediction time on a Nvidia Quadro M2000M is about \SI{178}{\micro\second} per step. This value can be further improved by optimizing the implementation. As one measurement cycle of an ultrasonic signal requires at least \SI{30}{\milli\second} due to propagation effects and the detection range that needs to be covered. Consequently, the proposed method is already suited well for real-time classification even if other algorithms need to run simultaneously on the in-car processing unit and the processing is low-performance.\\
If necessary, the processing effort and the amount of data to be fed to the network can be further reduced by using the absolute values of the envelope. This approach is of increased interest, as automotive ultrasonic sensors do usually not provide complex envelope data. By calculating the absolute value of the envelope the signal phase is lost. This information loss also affects the performance of the CapsNet. When applying the same network architecture as before and the same measurement samples, a validation accuracy of only 90.6\% is achieved. Consequently, the loss of information leads to the expected drop of the classification accuracy. Due to the performance degradation it is recommended to use complex input data if possible. However, the accuracy is still sufficient for ultrasonic height classification, especially if the classification is supported by a confirmation using subsequent measurements.\\
In addition to the CapsNet approaches, we tested our data using a conventional CNN. To create comparable results, we use the same datasets than for the CapsNets. Also the network architecture is very similar, as the first two layers are the same. After the second layer we use a conventional maximum pooling layer with a size of $2 \times 2$. The routing-by-agreement step is replaced by a dense layer with 64 neurons followed by an output layer that uses softmax activation. The cost function is a categorical crossentropy. \\
As before, the network is tested with complex input data first. The yielded validation accuracy is 98.9\%. The decrease of the accuracy  results from the pooling layer. In addition, the CNN does not preserve the amplitude locations so that an additional error is produced. The error in our dataset is only 0.7\%. The prediction time increased to \SI{213}{\micro\second} per step. Consequently the proposed CapsNet approach does not only provide better validation accuracy, but furthermore requires less computational effort. \\
In case of feeding the network with the absolute valued data, a decrease in the validation accuracy can be observed. The reached accuracy is only 87.8\%. Compared to the CapsNet approach but also to the complex CNN method, the degradation of the accuracy is significant. Although the calculation time is \SI{199}{\micro\second} and thus lower than in the complex CNN method, the accuracy is 2.8\% lower than in the absolute valued CapsNet approach. Even if subsequent measurements are used to confirm the classification height, the 
suitability of this approach for autonomous driving must be strictly evaluated.\\
The results of the CapsNet- and CNN-based classification are summarized in Table\:\ref{result_tab}. Considering both, accuracy and computational effort, the proposed CapsNet-based classification approach that uses complex input data shows the best performance. If absolute valued input data is available only, the CapsNet method also performs better than the CNN-based approach. As it can be seen from the table, our proposed method shows an outstanding performance for classification and is suited well for the integration in real-time ultrasonic systems for automotive scenarios due to the improved prediction time.
\begin{table}[tb]
\centering
\caption{Comparison of CapsNet- and CNN-based approaches}
\label{result_tab}
\begin{tabular}{l r r }
& Validation Accuracy & Prediction Time\\ \cline{1-3}
Complex CapsNet & 99.6\% & \SI{178}{\micro\second} \\ 
Absolute CapsNet & 90.6\% & \SI{170}{\micro\second} \\ 
Complex CNN & 98.9\% & \SI{213}{\micro\second} \\ 
Absolute CNN & 87.8\% & \SI{199}{\micro\second} \\ 
\end{tabular}
\end{table}


\section{Conclusion}
CapNets are an upcoming network architecture that can be used in a wide variety of applications. In the paper, we show how to apply this network type to ultrasonic data from automotive ultrasonic sensors to classify object height. In contrast to image-based classification the ultrasonic raw data does not allow an easy data interpretation by human. Thus we apply a special preprocessing on the ultrasonic data first. The preprocessing improves the signal quality and optimizes classification performance. Subsequently, we use a CapsNet with routing-by-agreement. By feeding the network with the complex envelope data of ultrasonic measurements, validation accuracies of more than 99\% are reached. A comparison to conventional CNN-based approaches additionally shows improved classification accuracy and reduced computational effort of CapsNets.\\
Object classification using automotive ultrasonic sensors is still at the beginning. Height classification is just one example where deep learning is applicable. There are lots of other object properties, such as width, material or speed, where classification algorithms can help to improve the functionality of ultrasonic sensors.\\
The limitation to a single ultrasonic sensor is also not necessary at all. Using a couple of sensors, e.g. in a front bumper, or subsequent measurements allows to determine further object properties and to increase the classification accuracy. \\
Altogether the potential of classification in automotive ultrasonic sensor technology is enormous. However, a lot of research and testing is required to evaluate the scope of ultrasonic object classification and to make ultrasonic technology suitable for autonomous driving.


\section{Acknowledgment}
For his constructive comments and input, we thank Martin Simon. Further we are grateful to Paul Rostocki for providing the ultrasonic sensors and Dr. Heinrich Gotzig for the technical supervising.


\bibliographystyle{IEEEtran}
\bibliography{IEEEabrv,biblio_traps_dynamics}

\begin{thebibliography}{10}
\providecommand{\url}[1]{#1}
\csname url@samestyle\endcsname
\providecommand{\newblock}{\relax}
\providecommand{\bibinfo}[2]{#2}
\providecommand{\BIBentrySTDinterwordspacing}{\spaceskip=0pt\relax}
\providecommand{\BIBentryALTinterwordstretchfactor}{4}
\providecommand{\BIBentryALTinterwordspacing}{\spaceskip=\fontdimen2\font plus
\BIBentryALTinterwordstretchfactor\fontdimen3\font minus
  \fontdimen4\font\relax}
\providecommand{\BIBforeignlanguage}[2]{{%
\expandafter\ifx\csname l@#1\endcsname\relax
\typeout{** WARNING: IEEEtran.bst: No hyphenation pattern has been}%
\typeout{** loaded for the language `#1'. Using the pattern for}%
\typeout{** the default language instead.}%
\else
\language=\csname l@#1\endcsname
\fi
#2}}
\providecommand{\BIBdecl}{\relax}
\BIBdecl

\bibitem{us_parking}
Lee,~Y. and Chang,~S., ``Development of a verification method on
  ultrasonic-based perpendicular parking assist system,'' in \emph{The 18th
  IEEE International Symposium on Consumer Electronics (ISCE 2014)}, June 2014,
  pp. 1--3.

\bibitem{advanced_us}
H.~Jeong,~S., G.~Choi,~C., and N.~Oh,~J., ``Low cost design of parallel parking
  assist system based on an ultrasonic sensor,'' vol.~11, pp. 409--416, 06
  2010.

\bibitem{prenatal_us}
Bozma,~H.~I. and Kemik,~E., ``Model-based multi-objective analysis of
  ultrasound image sequences in prenatal diagnosis,'' in \emph{Proceedings of
  IEEE Computer Society Conference on Computer Vision and Pattern Recognition},
  Jun 1997, pp. 942--947.

\bibitem{pattern_recog}
\BIBentryALTinterwordspacing
Wu,~W.-J., Lin,~S.-W., and Moon,~W.~K., ``Combining support vector machine with
  genetic algorithm to classify ultrasound breast tumor images,''
  \emph{Computerized Medical Imaging and Graphics}, vol.~36, no.~8, pp. 627 --
  633, 2012. [Online]. Available:
  \url{http://www.sciencedirect.com/science/article/pii/S0895611112001310}
\BIBentrySTDinterwordspacing

\bibitem{us_mlp}
Nugroho,~H.~A., Rahmawaty,~M., Triyani,~Y., Ardiyanto,~I., Choridah,~L., and
  Indrastuti,~R., ``Texture analysis and classification in ultrasound medical
  images for determining echo pattern characteristics,'' in \emph{2017 7th IEEE
  International Conference on System Engineering and Technology (ICSET)}, Oct
  2017, pp. 23--26.

\bibitem{us_cnn}
Yap,~M.~H., Pons,~G., Martí,~J., Ganau,~S., Sentís,~M., Zwiggelaar,~R.,
  Davison,~A.~K., and Martí,~R., ``Automated breast ultrasound lesions
  detection using convolutional neural networks,'' \emph{IEEE Journal of
  Biomedical and Health Informatics}, pp. 1--1, 2018.

\bibitem{us_rnn}
Yang,~X., Yu,~L., Wu,~L., Wang,~Y., Ni,~D., Qin,~J., and Heng,~P.-A.,
  ``Fine-grained recurrent neural networks for automatic prostate segmentation
  in ultrasound images,'' 2016.

\bibitem{us_segmentation}
Li,~Y.~D., ``Segmentation of medical ultrasound images using convolutional
  neural networks with noisy activating functions,'' 2016.

\bibitem{us_image_rec}
Byra,~M., Sznajder,~T., Korzinek,~D., Piotrzkowska-Wroblewska,~H.,
  Dobruch-Sobczak,~K., Nowicki,~A., and Marasek,~K., ``Impact of ultrasound
  image reconstruction method on breast lesion classification with neural
  transfer learning,'' 2018.

\bibitem{capsnet_paper1}
\BIBentryALTinterwordspacing
Sabour,~S., Frosst,~N., and Hinton,~G.~E., ``Dynamic routing between
  capsules,'' \emph{CoRR}, vol. abs/1710.09829, 2017. [Online]. Available:
  \url{http://arxiv.org/abs/1710.09829}
\BIBentrySTDinterwordspacing

\bibitem{capsnet_paper2}
\BIBentryALTinterwordspacing
Hinton,~G.~E., Sabour,~S., and Frosst,~N., ``Matrix capsules with {EM}
  routing,'' in \emph{International Conference on Learning Representations},
  2018. [Online]. Available: \url{https://openreview.net/forum?id=HJWLfGWRb}
\BIBentrySTDinterwordspacing

\bibitem{digital_comm}
\BIBentryALTinterwordspacing
Proakis,~J., \emph{Digital Communications}, ser. Electrical engineering
  series.\hskip 1em plus 0.5em minus 0.4em\relax McGraw-Hill, 2001. [Online].
  Available: \url{https://books.google.de/books?id=sbr8QwAACAAJ}
\BIBentrySTDinterwordspacing

\end{thebibliography}

\end{document}